\documentclass[11pt]{article}

\usepackage{amsmath,amsfonts,amsthm,amssymb,color}
\usepackage{mathrsfs}
\usepackage{fancybox}
\usepackage{tikz}
\usepackage{appendix}
\usepackage{enumerate}
\usepackage{bm}
\usepackage{subeqnarray}
\usepackage{cases}
\usepackage{pifont}
\usepackage{graphicx}
\usepackage{placeins}
\usepackage{epstopdf}
\usepackage{authblk}
\usepackage[procnumbered,ruled,boxed,linesnumbered]{algorithm2e}
\DontPrintSemicolon
\SetKw{KwAnd}{and}
\SetProcNameSty{textsc}
\SetFuncSty{textsc}
\SetKwRepeat{Do}{do}{while}
\usepackage[nocompress]{cite}
\usepackage[colorlinks=true,linkcolor=red,citecolor = blue]{hyperref}

\usepackage{hyperref}
\usepackage{cleveref} %
\usepackage{thm-restate}
\crefformat{equation}{(#2#1#3)}

\usepackage{caption}
\usepackage{bbm}
\usepackage{tablefootnote}

\SetAlCapSkip{.5em}

\usepackage[margin=1.0in]{geometry}

\newtheorem{theorem}{Theorem}%

\theoremstyle{definition}

\newenvironment{fminipage}%
{\begin{Sbox}\begin{minipage}}%
		{\end{minipage}\end{Sbox}\fbox{\TheSbox}}

\def\eq#1{\begin{equation*}\begin{split}#1\end{split}\end{equation*}}

\def\eql#1#2{\begin{equation}{#1}\begin{split}#2\end{split}\end{equation}}

\def\pr#1{\left( #1 \right ) }
\def\br#1{\left[ #1 \right ] }
\def\dr#1{\left\{#1\right\}}

\def\calD{\mathcal{D}}

\def\calF{\mathcal{F}}

\def\calJ{\mathcal{J}}
\def\calL{\mathcal{L}}

\def\calS{\mathcal{S}}

\newcommand \eps{\epsilon}

\newcommand \Real{\mathbb{R}}

\newcommand\na{\nabla}

\newcommand\x{\times}

\def\prb#1{\big( #1 \big)}
\def\prB#1{\Big( #1 \Big)}

\SetKwInput{KwRequire}{Require}

\def\KL#1#2{D_{\mathrm{KL}}(#1 \,\Vert\, #2)}

\newcommand\Eb{\mathbb{E}}
\newcommand\wt{\omega}
\newcommand\adv{\hat{A}}
\newcommand\stopgrad{\texttt{sg}}

\newcommand\tpi{\pi}
\newcommand\cpi{\mathbb{P}_{\pi}}
\newcommand\cmu{\mathbb{P}_{\mu}}
\def\cpx#1{\mathbb{P}_{#1}}
\newcommand\policyspace{\Pi}

\title{Bellman Policy Optimization}

\author{%
Zhuoqing Song\textsuperscript{1,2}, Haotian Xu\textsuperscript{1},
Xikun Zhang\textsuperscript{1}, 
and Lidong Bing\textsuperscript{1}\\
\textsuperscript{1}Apodex US, Inc.\quad
\textsuperscript{2}Princeton University
}

\date{}

\begin{document}

\maketitle

\begin{abstract}
Reinforcement learning with verifiable rewards (RLVR) improves the reasoning capabilities of large language models (LLMs).
We introduce Bellman Policy Optimization (BPO), a critic-free method derived from Policy Mirror Descent (PMD).
For autoregressive generation with terminal rewards, BPO uses the Bellman equations to reformulate PMD as a trajectory-level objective.
The reformulation avoids estimating state values at intermediate states. We prove that it has the same unique optimal solution as the original PMD objective.
We derive the practical BPO loss by approximating this objective. Its mismatch-correction weight is a smoothed ratio of complementary token probabilities.       
Experiments on mathematical reasoning benchmarks demonstrate the effectiveness of BPO.
\end{abstract}

\section{Introduction}
Reinforcement learning with verifiable rewards (RLVR) has become an important approach for improving the reasoning capabilities of large language models (LLMs)~\cite{shao2024deepseekmath,guo2025deepseek,yang2025qwen3,yu2026dapo,abdin2025phi,wen2026reinforcement}.
In RLVR, models are trained using outcome-level rewards assigned to generated responses by task-specific verifiers.
Recent work has shown that choices in policy optimization can have a substantial impact on both reasoning performance and training stability~\cite{yu2026dapo,zheng2025gspo,qi2026rethinking,ma2026fipo,mao2026beyond,kang2026vimpo}.

\begin{figure}[t]
\centering
\includegraphics[width=0.8\linewidth]{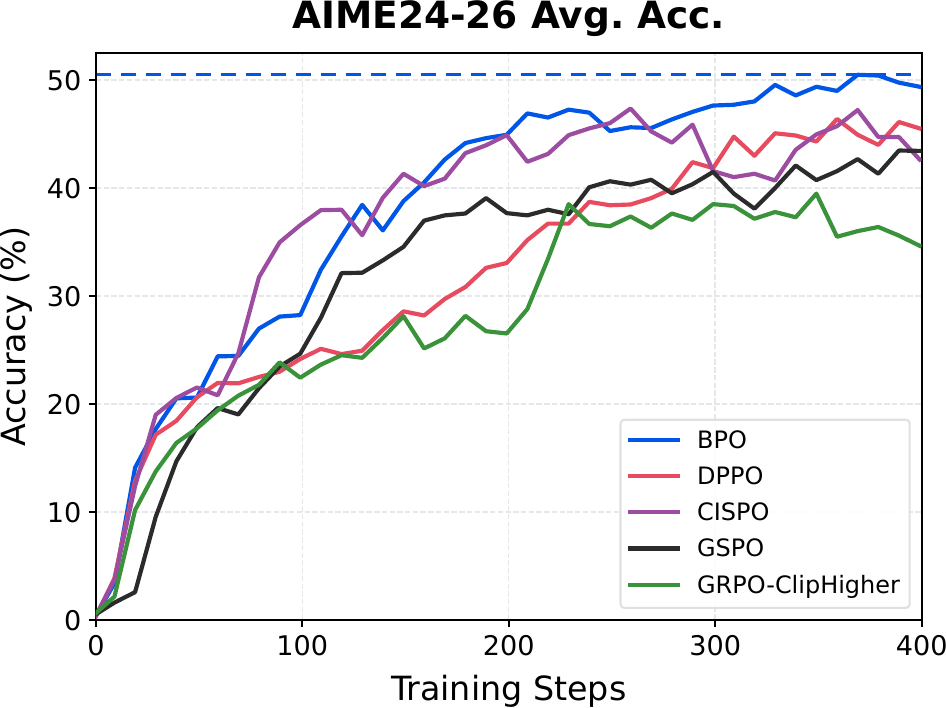}
\caption{Evaluation accuracy during training on Qwen3-30B-A3B-Base, comparing BPO with GRPO-ClipHigher, GSPO, CISPO, and DPPO. All methods are trained for 400 training steps on the English subset of DAPO-Math-17k under identical experimental settings except for the policy loss. Each rollout batch contains 256 prompts, with a group size of 16 responses per prompt and a maximum response length of 16384 tokens. One training step consists of one rollout batch followed by eight optimizer updates. Curves show mean Avg@32 accuracy across AIME24--26, with 32 responses sampled per question to estimate Pass@1. The horizontal dashed line marks BPO's peak mean accuracy.}
\label{fig:mainaime}
\end{figure}

Group-Relative Policy Optimization (GRPO) and its variants are widely used in RLVR~\cite{shao2024deepseekmath,yu2026dapo,liu2025understanding,zheng2025gspo,chen2025minimax,zhang2026design,qi2026rethinking,liu2026prorl}. GRPO samples multiple responses for each prompt and computes advantages by normalizing their rewards within the group. Under outcome supervision, all tokens in a response share the same advantage. GRPO uses these advantages in a PPO-style clipped surrogate objective with token-level importance-sampling ratios~\cite{schulman2017proximal}. This formulation and its variants avoid training a value model and support large-scale reasoning-model training~\cite{guo2025deepseek,yang2025qwen3,abdin2025phi,chen2025minimax,blakeman2025nvidia,blakeman2026nemotron,team2025every}.  

We start from Policy Mirror Descent (PMD)~\cite{lan2023policy,xiao2022convergence}. PMD updates the policy using action values and a divergence penalty. However, directly applying this update to language generation requires value estimates at intermediate states. A common approach is to train a separate value model, which increases memory and computational costs~\cite{shao2024deepseekmath}. Learned value estimates can also be inaccurate on reasoning tasks~\cite{pmlr-v267-kazemnejad25a}. We seek a reformulation of PMD that recovers the same policy update without training a value model.

We introduce Bellman Policy Optimization (BPO), a critic-free policy optimization method derived from PMD. Our derivation builds on prior work that reparameterizes rewards and values using policy likelihood ratios~\cite{rafailov2023direct,zeng2024token,rafailovr,kang2026vimpo}. For autoregressive generation with terminal rewards, the Bellman equations express each token advantage as a difference between value functions at consecutive states. These differences telescope along a response, leaving only the terminal reward and the initial value. We combine this identity with the PMD optimality condition to derive a trajectory-level objective. 
The initial value is the expected reward for a prompt and can be estimated from sampled responses. The reformulation therefore avoids estimating values or advantages at intermediate states.
We prove that this objective and the original PMD objective have the same unique optimal solution on states reachable under the rollout policy. 

We then derive the practical BPO loss through a sequence of approximations. We estimate the initial value using the mean reward of responses sampled for each prompt. We approximate the gradient of the trajectory-level objective and replace the full KL divergence with a binary approximation~\cite{qi2026rethinking}. The resulting gradient includes a mismatch-correction weight determined by the rollout and current token probabilities. We apply additive smoothing to this weight for numerical stability. In the practical BPO loss, this weight replaces the importance-sampling ratio used in GRPO. 

We evaluate BPO on Qwen3-30B-A3B-Base trained on DAPO-Math-17k.
We compare against GRPO-ClipHigher~\cite{shao2024deepseekmath,yu2026dapo}, GSPO~\cite{zheng2025gspo}, CISPO~\cite{chen2025minimax}, and DPPO~\cite{qi2026rethinking} under identical experimental settings.
As shown in Figure~\ref{fig:mainaime}, BPO achieves a peak average accuracy of $50.5\%$ across AIME 2024--2026~\cite{aimeproblems}.
The gains over these four baselines are $11.0$, $7.0$, $3.1$, and $4.1$ percentage points, respectively.

Our main contributions are:

\begin{itemize}
\item We derive a critic-free, trajectory-level reformulation of PMD for autoregressive generation with terminal rewards, avoiding value or advantage estimation at intermediate states. We prove that this reformulation and the original PMD objective have the same unique optimal solution on states reachable under the rollout policy.

\item We derive the practical BPO loss from this reformulation through group-based estimation and gradient approximations. The loss replaces GRPO's importance-sampling ratio with a mismatch-correction weight given by a smoothed ratio of complementary token probabilities.

\item We evaluate BPO on mathematical reasoning benchmarks using Qwen3-30B-A3B-Base. BPO achieves a peak average accuracy of $50.5\%$ across AIME 2024--2026, outperforming GRPO-ClipHigher, GSPO, CISPO, and DPPO by $3.1$--$11.0$ percentage points.

\end{itemize}

\FloatBarrier

\section{Preliminaries}\label{sec:prel1}

\subsection{Notation}\label{sec:prel2}  
Let $\mathcal{V}$ denote the vocabulary. 
Let $x$ denote a prompt and let $y=(y_1,\ldots,y_T)$ denote a response. 
Both $x$ and $y$ are token sequences over $\mathcal{V}$. 
In response $y$, $y_t$ is the $t$-th token and the length $T$ may vary across responses. 
We also use $|y|$ to denote its length.
 
For $1\leq t\leq u\leq T$, let $y_{t:u}=(y_t,\ldots,y_u)$ denote the subsequence of $y$ from positions $t$ through $u$. We write $y_{<t} = y_{1:t-1}$, $y_{\leq t} = y_{1:t}$, and $y_{\geq t} = y_{t:T}$.   

For any token sequences $z_1,\ldots,z_k$, $(z_1,\ldots,z_k)$ denotes their concatenation. For instance, $(x,y_{<t})$ denotes the prompt $x$ followed by the response subsequence $y_{<t}$. 
For a function whose argument is a token sequence, we write $f(z_1,\ldots,z_k)$ as shorthand for $f((z_1,\ldots,z_k))$ whenever no ambiguity arises. 

\subsection{RLVR as an Episodic MDP}
Reinforcement learning with verifiable rewards (RLVR) for large language models (LLMs) can be formulated as a finite-horizon episodic Markov Decision Process (MDP). 

Let $D$ denote a prompt set and let $\calD$ denote a prompt distribution on $D$. 
We assume $\calD$ is uniform on $D$ for simplicity. The analysis extends directly to more general prompt distributions.

\paragraph{States and transitions.}
For a prompt $x\in D$, the initial state $s_1$ is defined as the sequence $x$ itself. 
At step $t>1$, the state $s_t$ concatenates the prompt $x$ and the generated prefix $y_{<t}$:  
$
    s_t = (x,y_{<t}).
$  
The transition is deterministic: appending token $y_t$ to
$s_t=(x,y_{<t})$ yields the next state
$s_{t+1}=(x,y_{\leq t})$.  

\paragraph{Policies and completion distributions.}
Let $\Delta(\mathcal{V}) = \dr{p: \mathcal{V}\rightarrow[0,1]| \sum_{v\in\mathcal{V}}p(v)=1}$ denote the set of probability distributions on $\mathcal{V}$. 
A stochastic policy is a mapping $\tpi:\calS \rightarrow \Delta(\mathcal{V})$. In particular, $\tpi(\cdot | x,y_{<t})\in\Delta(\mathcal{V})$ denotes the next-token distribution at position $t$. 
The policy space is denoted by $\policyspace = \Delta(\mathcal{V})^{\calS}$. 
For any policy $\pi$, let $\Pi_{\pi} \subseteq \Pi$ denote the set of policies that are absolutely continuous with respect to $\pi$.

For any policy $\tpi\in \policyspace$ and state $s_t\in \calS$, we use $\mathbb{P}_{\pi}(\cdot | s_t)$ to denote the completion distribution induced by $\tpi$, i.e., $\mathbb{P}_{\pi}(y_{\geq t}| s_t) = \prod_{u=t}^{|y|}\pi(y_u|s_u)$, where $y_{\geq t} = y_{t:|y|}$ as mentioned in Section~\ref{sec:prel2}. 

The action at step $t$ is the $t$-th token $y_t$ in the generated response, and is sampled from the policy $\tpi(\cdot | x, y_{<t})$. 
In practice, $\pi$ is parameterized by an autoregressive LLM. 
The response $y = (y_1, \dots, y_T)$ terminates when the EOS token is generated or the response length limit $T_{\max}$ is reached. 

\paragraph{Rewards and objective.}
A deterministic verifier $R$ assigns the terminal reward. 
Without loss of generality, we assume $|R(x, y)| \leq 1$ for any prompt-response pair $(x, y)$.   

The policy objective is  
\eql{\label{eq:J}}{
    \calJ(\pi)
    =
    \mathbb{E}_{x\sim \mathcal{D}, y\sim \cpi(\cdot| x)}
    \left[
    R(x,y)
    \right].
}

\subsection{Group-Relative Policy Optimization}
In practical RLVR, responses are generated by a rollout policy $\mu$, while policy optimization is performed on the current policy $\pi$.
Policy updates between rollout generation and training steps, together with numerical discrepancies between the rollout and training engines, can cause the two policies to differ~\cite{zheng2025stabilizing,qi2025defeating,qi2026rethinking,yao2026rethinking,yao2025rollout}.
Group-Relative Policy Optimization (GRPO)~\cite{shao2024deepseekmath} is a critic-free, PPO-style policy optimization method~\cite{schulman2017proximal}.
For each prompt $x \sim \calD$, GRPO independently samples a group of $G$ responses $\dr{y^i}_{i=1}^{G}$ from $\cmu(\cdot | x)$.
Let $R_i = R(x, y^i)$ denote the reward of response $y^i$.
GRPO assigns each response the group-normalized advantage
\eql{\label{eq:adv}}{
    \adv^i = \frac{R_i - \texttt{mean}\prb{\{R_j\}_{j=1}^{G}}}{\texttt{std}\prb{\{R_j\}_{j=1}^{G}}}.
}

GRPO applies PPO-style clipping to the token-level importance ratio.
For the $t$-th token in response $y^i$, its per-token loss is defined as
\eql{\label{eq:grpoloss}}{
    \calL_{i,t}^{\mathrm{GRPO}}(\pi) = -\min\dr{r_t^i\adv^i,\; \mathrm{clip}\pr{r_t^i, 1-\eps_{\mathrm{low}}, 1+\eps_{\mathrm{high}}}\adv^i},
}
where $r_t^i$ is the token-level importance ratio defined as
\eql{\label{eq:isratio}}{
    r_t^i = \frac{\pi(y_t^i | s_t^i)}{\mu(y_t^i | s_t^i)}.
}

The following per-token loss has the same gradient as~\eqref{eq:grpoloss} and is therefore equivalent for gradient-based training:  
\eql{\label{eq:grpoloss-log}}{
    \widetilde{\calL}_{i,t}^{\mathrm{GRPO}}(\pi) = -\adv^i M_t^i \stopgrad\pr{r_t^i}\log\pi(y_t^i | s_t^i),
}
where the clipping mask $M_t^i$ is defined as
\eql{\label{eq:maskgrpo1}}{
    M_t^i = \left\{\begin{array}{ll} 0, & \adv^i > 0 \text{ and } r_t^i > 1+\eps_{\mathrm{high}}, \\ 0, & \adv^i < 0 \text{ and } r_t^i < 1-\eps_{\mathrm{low}}, \\ 1, & \text{otherwise.} \end{array}\right.
}

\subsection{Bellman Equations with Terminal Rewards}\label{sec:BellmanEquations1}  
We next introduce value functions and Bellman equations~\cite{sutton1998reinforcement} for the terminal-reward setting described in Section~\ref{sec:prel1}.  
For a fixed policy $\tpi$ and state $s_t = (x, y_{<t})$, define the value function as the expected reward under policy $\tpi$ given state $s_t$, i.e., 
\eql{\label{eq:V1}}{
    V^\tpi(s_t)
    =
    \Eb_{z \sim \cpi(\cdot|s_t)}\left[R(x,(y_{<t}, z)) \right],
}
where the expectation is taken over all completions under policy $\tpi$ given state $s_t = (x, y_{<t})$. 
The initial value $V^{\tpi}(s_1) = V^{\tpi}(x)$ is therefore the expected reward for prompt $x$ under policy $\pi$.  

Since intermediate rewards are zero, for a non-terminal state $s_t$, the action-value function is defined as
\eql{\label{eq:Q2}}{
    Q^\tpi(s_t,y_t)
    =
    \mathbb{E}_{s'\sim \mathbb{P}(\cdot| s_t,y_t)}
    \left[V^\tpi(s')\right] 
    = V^{\tpi}(s_{t+1}),
}
where $s_{t+1} = (s_t, y_t)$ and the second equality follows from the deterministic transition described in Section~\ref{sec:prel1}.

For a non-terminal state $s_t$, the advantage function is defined as  
\eql{\label{eq:A1}}{
    A^\tpi(s_t,y_t) = Q^\tpi(s_t,y_t) - V^\tpi(s_t).
}
The advantage function $A^\tpi(s_t,y_t)$ measures how favorable the current action $y_t$ is at state $s_t$ relative to the expected action value under policy $\tpi(\cdot | s_t)$. 
It follows from~\eqref{eq:V1} and~\eqref{eq:A1} that 
\eql{\label{eq:avgA1}}{
    \sum_{y_t \in \mathcal{V}} \tpi(y_t | s_t) A^{\tpi}(s_t, y_t) = 0.  
} 

For any non-terminal state $s_t$, the Bellman equation is
\eq{
    V^\tpi(s_t)
    =
    \sum_{y_t'\in \mathcal{V}} \tpi(y_t'| s_t) Q^\tpi(s_t,y_t')
    =
    \sum_{s'=(s_t,y_t'): y_t'\in \mathcal{V}} \tpi(y_t' | s_t) V^{\tpi}(s').
}
For a complete response $y$, the terminal state is $s_{|y|+1} = (x, y)$. The terminal condition is
\eq{
    V^{\tpi}(s_{|y| + 1}) = R(x,y).
}

\subsection{Policy Mirror Descent}\label{sec:PMD} 
Let $\calS$ denote the state space, and let $\mathcal{V}$ denote the vocabulary as in Section~\ref{sec:prel1}.   
Given a real-valued function $f: \mathcal{S} \x \mathcal{V} \rightarrow \Real$, Policy Mirror Descent (PMD)~\cite{geist2019theory,lan2023policy,xiao2022convergence,tomar2020mirror} solves the following optimization problem for each state $s \in \calS$:
\eql{\label{eq:PMD}}{
    \max_{\pi(\cdot | s)\in \Delta(\mathcal{V})} \quad \Eb_{y\sim \pi(\cdot| s)}\br{f(s, y)} - \frac{1}{\eta} {\KL{\pi(\cdot| s)}{\mu(\cdot | s)}},                            
}
where $\mu$ is the behavior policy (rollout policy) and $\eta > 0$ is the step size. 

The optimization problem~\eqref{eq:PMD} has the unique solution
\eql{\label{eq:piexpw}}{
    \pi^+_{\mu, f}(y|s) = \frac{\mu(y|s)\exp\pr{\eta f(s, y)}}{Z_{\mu, f}(s) },
}
where the partition function is
\eq{
    Z_{\mu, f}(s) = \Eb_{y'\sim \mu(\cdot|s)} \exp\pr{\eta f(s, y')}.  
}

\subsection{Binary KL Divergence}\label{sec:binary_kl}
Let $\pi, \mu \in \Pi$ be two policies on $\mathcal{V}$. 
For a fixed state $s$, their KL divergence is
\eq{
    \KL{\pi(\cdot|s)}{\mu(\cdot|s)}
    =
    \sum_{y\in\mathcal{V}}
    \pi(y | s)
    \log
    \frac{\pi(y | s)}{\mu(y | s)}.
}
Computing the full KL divergence requires logits for the entire vocabulary and can be costly.

The binary KL divergence introduced in~\cite{qi2026rethinking} approximates the full KL divergence. 
For a given action $y\in\mathcal{V}$, it partitions the action space into $\dr{y}$ and
its complement $\mathcal{V}\setminus\dr{y}$. 
This induces the following Bernoulli distributions:
\eq{
    \pi^{\mathrm{bin}}_y(\cdot| s)
    = \pr{\pi(y| s), 1 - \pi(y| s)},
    \quad
    \mu^{\mathrm{bin}}_y(\cdot| s)
    = \pr{\mu(y| s), 1 - \mu(y| s)}.
}
We define the binary KL divergence associated with action $y$ as
\eql{\label{eq:binary_kl}}{
    &\quad D^{\mathrm{bin}}_{\mathrm{KL}}
    \pr{\pi(\cdot| s)\,\Vert\,\mu(\cdot| s); y}
    =
    D_{\mathrm{KL}}
    \pr{\pi^{\mathrm{bin}}_y(\cdot| s) \,\Vert\, \mu^{\mathrm{bin}}_y(\cdot| s)}
    \\
    &= \pi(y | s)\log\frac{\pi(y | s)}{\mu(y | s)} + \pr{1-\pi(y | s)}\log\frac{1-\pi(y | s)}{1-\mu(y | s)}.
}
As shown in~\cite{qi2026rethinking}, binary KL divergence provides a lower bound for KL divergence, i.e., 
\eq{
    0 \leq D^{\mathrm{bin}}_{\mathrm{KL}}\pr{\pi(\cdot| s) \,\Vert\, \mu(\cdot| s); y}
    \leq D_{\mathrm{KL}}\pr{\pi(\cdot| s) \,\Vert\, \mu(\cdot| s)}, \quad \forall y \in \mathcal{V}.  
} 

\section{Bellman Policy Optimization}
We first define the BPO loss and then derive it from PMD.

\paragraph{BPO Loss Definition.}
For each prompt $x\sim\calD$, we independently sample a group of $G$ responses $\dr{y^i}_{i=1}^G$.   
For the $t$-th token of response $y^i$, the BPO loss is
\eql{\label{eq:bpoloss1}}{
    \calL^{\text{BPO}}(\pi) = - 
    \adv^i M^i_t\min\dr{\stopgrad\pr{\wt^i_{t}}, C} \log\pi(y^i_t| x, y^i_{<t}),  
}  
where $\adv^i$ is the advantage of response $y^i$ in the group $\dr{y^i}_{i=1}^G$ defined in~\eqref{eq:adv} and the terms $\wt^i_t$, $C$, and $M^i_t$ are defined below:
\begin{itemize}
    \item $\wt^i_t$ is the mismatch-correction weight for the $t$-th token in response $y^i$, defined as  
    \eql{\label{eq:wti1}}{
        \wt^i_t = \frac{1 + \eps - \mu(y^i_t|x, y^i_{<t})}{1 + \eps - \pi(y^i_t|x, y^i_{<t})},  
    }

    \item The constant $C$ caps $\wt^i_t$ to stabilize training. The mask $M_t^i$ follows the GRPO clipping rule in~\eqref{eq:maskgrpo1}, with the weight $\wt_t^i$ replacing the importance-sampling ratio: 
    \eq{
        M^i_t = 
        \left\{
        \begin{array}{ll}
            0, & \adv^i > 0 \text{ and } \wt^i_t > 1 + \eps_{\text{high}} \\
            0, & \adv^i < 0 \text{ and } \wt^i_t < 1 - \eps_{\text{low}} \\
            1, & \text{otherwise.}
        \end{array}
        \right.
    }
\end{itemize}
The BPO per-token loss retains the GRPO form in~\eqref{eq:grpoloss-log} and replaces the importance-sampling ratio $r_t^i$ with the truncated mismatch-correction weight $\min\dr{\stopgrad\pr{\wt_t^i}, C}$.

\paragraph{Derivation of the BPO Objective.} 
We derive the BPO objective~\eqref{eq:bpoloss1} in three steps:
\begin{itemize}
    \item Step 1 (Starting Point,~\eqref{eq:PMDadv1}): we instantiate PMD with $f(s, y) = A^{\mu}(s, y)$, where $A^{\mu}(s, y)$ is the advantage function introduced in Section~\ref{sec:BellmanEquations1}.
    \item Step 2 (Critic-free PMD, Section~\ref{sec:criticfreePMD}): we derive a critic-free objective with the same optimal solution as advantage-based PMD. This avoids estimating $A^\mu(s,y)$ at individual states.  
    \item Step 3 (Practical Approximation, Section~\ref{sec:binaryKL1}): we approximate the critic-free PMD objective from Step~2 to obtain the BPO objective~\eqref{eq:bpoloss1}.  
\end{itemize}

\paragraph{Starting Point: Advantage-Based PMD.}
We first instantiate the general PMD objective in~\eqref{eq:PMD} with
the rollout-policy advantage $A^{\mu}(s, y)$ of the terminal-reward MDP in Section~\ref{sec:BellmanEquations1}, i.e., 
$
    f(s,y)=A^\mu(s,y).  
$
Classical PMD is commonly formulated using the action-value function
$Q^\mu(s_t, y_t)$~\cite{geist2019theory,lan2023policy,xiao2022convergence,tomar2020mirror}. Using $A^\mu$ gives the same update because
$
    Q^\mu(s_t, y_t)=A^\mu(s_t, y_t)+V^\mu(s_t),
$
and the term $V^\mu(s_t)$ does not depend on the action $y_t$. 

Advantage-based PMD solves the following problem for each state $s_t = (x, y_{<t})$:
\eql{\label{eq:PMDadv1}}{
    \max_{\pi(\cdot | s_t)\in \Delta(\mathcal{V})} \quad \Eb_{y_t\sim \pi(\cdot| s_t)}\br{A^{\mu}(s_t, y_t)} - \frac{1}{\eta} {\KL{\pi(\cdot| s_t)}{\mu(\cdot | s_t)}}.                       
} 
As discussed in Section~\ref{sec:PMD}, the optimization problem~\eqref{eq:PMDadv1} has the unique solution $\pi^+$, given at each state $s_t = (x, y_{<t})$ by 
\eql{\label{eq:piexpwadv1}}{
    \pi^+(y_t|s_t) = \frac{\mu(y_t|s_t)\exp\pr{\eta A^{\mu}(s_t, y_t)}}{Z_{\mu}(s_t) },
}
where the partition function is
\eq{
    Z_{\mu}(s_t) = \Eb_{y'_t\sim \mu(\cdot|s_t)} \exp\pr{\eta A^{\mu}(s_t, y'_t)}.
}
Here, we use $\pi^+$ and $Z_{\mu}$ as abbreviations for $\pi^+_{\mu, A^{\mu}}$ and $Z_{\mu, A^{\mu}}$.  

\subsection{Critic-Free Reformulation of Policy Mirror Descent}\label{sec:criticfreePMD}
Directly implementing the update~\eqref{eq:PMDadv1} requires estimating $A^\mu(s_t,y_t)$ at every visited intermediate state.  
This typically requires a learned critic or costly conditional rollouts.
We derive the critic-free reformulation in~\eqref{eq:delta2} by expressing rewards and values through policy likelihood ratios and the Bellman equations. This follows prior work on direct alignment~\cite{rafailov2023direct,zeng2024token,rafailovr,kang2026vimpo}. 

\paragraph{Critic-Free Reformulation of PMD.}
Let $\mu$ denote the rollout policy.
We assign each prompt $x\in D$ a fixed positive weight $\phi(x)$.
We will specify this weight in Section~\ref{sec:binaryKL1} to obtain the practical BPO loss.
We consider the following critic-free objective:
\eql{\label{eq:delta2}}{
    \min_{\pi\in\Pi_\mu}
    \calL(\pi) = 
    \Eb_{x\sim \calD, y\sim\cmu(\cdot|x)}\br{\phi(x) \cdot \frac{\delta(x,y;\pi,\mu)^2}{2 \eta             }},   
}
where the trajectory-level residual $\delta(x, y; \pi, \mu)$ is defined as
\eql{\label{eq:deltadef1}}{
    \delta(x, y; \pi, \mu) = \eta \prb{R(x, y) - V^{\mu}(x)} - \sum_{t=1}^{|y|} \prB{\log\frac{\pi(y_t | s_t)}{\mu(y_t | s_t)} + \KL{\mu(\cdot | s_t)}{\pi(\cdot | s_t)}}.   
}

The reformulation~\eqref{eq:delta2} does not depend on state-value functions at intermediate states. Instead, it only requires the terminal reward $R(x,y)$ and the initial value function $V^\mu(x)$.
The terminal reward $R(x,y)$ is directly provided by the verifier, while
$
    V^\mu(x) = \Eb_{y\sim\cmu(\cdot|x)}\br{R(x,y)}
$
is the expected reward for prompt $x$ under the rollout policy $\mu$.
We can estimate this expectation from rollout data.

\paragraph{Equivalence to PMD.}
Theorem~\ref{thm:delta2} shows that the critic-free objective~\eqref{eq:delta2} and PMD~\eqref{eq:PMDadv1} have the same unique optimal solution.
This equivalence holds for any positive weighting function $\phi$.
We specify $\phi$ in Section~\ref{sec:binaryKL1} when deriving the practical BPO loss~\eqref{eq:bpoloss1}.

\begin{theorem}\label{thm:delta2}
    Let the policy $\pi^+$ in~\eqref{eq:piexpwadv1} denote the optimal solution of the PMD problem~\eqref{eq:PMDadv1}. 
    Let $\phi: D\rightarrow\Real_+$ be any positive weight function on the prompt set $D$. 
    For any prompt $x \in D$, we solve the optimization problem~\eqref{eq:delta2}.  
    Then, this optimization problem admits a unique optimal solution $\hat{\pi}^+$ which equals $\pi^+$.  
    Here, uniqueness and equivalence are in the sense that
    any policy $\hat{\pi}^+\in \Pi_{\mu}$ that solves~\eqref{eq:delta2} induces the same completion distribution as $\pi^+$ for every prompt $x\in D$, i.e., $\cpx{\hat{\pi}^+}(\cdot|x) = \cpx{\pi^+}(\cdot|x)$.          
\end{theorem}

We now derive objective~\eqref{eq:delta2} and outline the proof of Theorem~\ref{thm:delta2}. 

\paragraph{Derivation and Proof of Sufficiency.}
We first explain how we derive the critic-free objective~\eqref{eq:delta2} from the original advantage-based PMD~\eqref{eq:PMDadv1} and show that the PMD solution $\pi^+$ minimizes the reformulated objective~\eqref{eq:delta2}.

\vspace{0.2cm}
\noindent$\bullet$~Step~1: We rewrite~\eqref{eq:piexpwadv1} as the PMD optimality condition and eliminate the state-dependent partition function $Z_{\mu}(s_t)$.  
For any prompt $x$, response $y$, and step $t$ with $1\leq t \leq |y|$, we can rewrite~\eqref{eq:piexpwadv1} as  
\eql{\label{eq:PMD2}}{
    \log\pi^+(y_t | s_t) - \log\mu(y_t | s_t) - \eta A^{\mu}(s_t, y_t) + \log Z_{\mu}(s_t) = 0.  
} 
Taking the expectation over $y_t \sim \mu(\cdot | s_t)$ and using~\eqref{eq:avgA1} gives  
\eql{\label{eq:logparti1}}{
    \log Z_{\mu}(s_t) = \KL{\mu(\cdot | s_t)}{\pi^+(\cdot | s_t)} + \eta \sum_{y_t\in \mathcal{V}} \mu(y_t | s_t) A^{\mu}(s_t, y_t) = \KL{\mu(\cdot | s_t)}{\pi^+(\cdot | s_t)}.      
} 
Then, substituting~\eqref{eq:logparti1} into~\eqref{eq:PMD2} yields 
\eql{\label{eq:PMDKL1}}{
    \log\pi^+(y_t | s_t) - \log\mu(y_t | s_t) - \eta A^{\mu}(s_t, y_t) + \KL{\mu(\cdot | s_t)}{\pi^+(\cdot | s_t)} = 0.  
}  

\noindent$\bullet$~Step 2: We next express the cumulative token-level advantages in terms of the terminal reward. By the Bellman equation, substituting~\eqref{eq:Q2} into~\eqref{eq:A1} gives
\eq{
    A^{\mu}(s_t,y_t)=V^{\mu}(s_{t+1}) - V^{\mu}(s_t).
}
Summing over the trajectory gives
\eql{\label{eq:sumA1}}{
    \sum_{t=1}^{|y|} A^{\mu}(s_t, y_t) = V^{\mu}(s_{|y|+1}) - V^{\mu}(s_1) = R(x, y) - V^{\mu}(x).   
} 

\noindent$\bullet$~Step~3: For any prompt-response pair $(x, y)$, summing~\eqref{eq:PMDKL1} over the trajectory and using~\eqref{eq:sumA1} gives 
\eq{
    \sum_{t=1}^{|y|} \pr{\log\pi^+(y_t | s_t) - \log\mu(y_t | s_t) + \KL{\mu(\cdot | s_t)}{\pi^+(\cdot | s_t)}} = \eta \pr{R(x, y) - V^{\mu}(x)},     
}
i.e.,  
\eql{\label{eq:sumdelta2}}{
    \delta(x, y; \pi^+, \mu) = 0. 
}
Minimizing the expected squared residual under the rollout distribution gives~\eqref{eq:delta2}. 

\paragraph{Proof Sketch of Necessity.}
It remains to show that every optimal solution of~\eqref{eq:delta2}
coincides with $\pi^+$ on all states reachable under $\mu$.
Define the token-level residual
\eq{
    d_\pi(s_t, y_t) = \eta A^\mu(s_t,y_t) -
    \prB{\log\pi(y_t|s_t) - \log\mu(y_t|s_t) + \KL{\mu(\cdot|s_t)}{\pi(\cdot|s_t)}}.  
}
For any feasible policy $\pi$ for which the above quantities are finite,
its conditional expectation under the rollout policy is zero:
\eq{
    \Eb_{y_t\sim\mu(\cdot|s_t)}\br{d_\pi(s_t,y_t)} = \eta\sum_{y_t\in\mathcal V}\mu(y_t|s_t)A^\mu(s_t,y_t) - \sum_{y_t\in\mathcal V}\mu(y_t|s_t)\log\frac{\pi(y_t|s_t)}{\mu(y_t|s_t)} - \KL{\mu(\cdot|s_t)}{\pi(\cdot|s_t)} = 0.  
}
Therefore, the cumulative token-level residuals form a martingale
under the rollout policy $\mu$.
For any optimal solution of~\eqref{eq:delta2}, the terminal value of this martingale equals the trajectory-level residual and is zero almost surely. Since the response length is bounded by $T_{\max}$, the martingale property then implies that every partial sum is zero almost surely. Thus, each token-level residual vanishes.
The complete proof is provided in Appendix~\ref{sec:prfdelta2}. 

\subsection{Practical Approximation}\label{sec:binaryKL1}

We obtain the practical BPO loss~\eqref{eq:bpoloss1} by approximating the critic-free objective~\eqref{eq:delta2}. We linearize the squared-residual objective and estimate the initial value and prompt-dependent scaling factor from grouped rollouts. We then approximate the full reverse KL divergence using binary KL divergence and apply smoothing, masking, and clipping. 

For response $y^i$, define the response-level loss
$\calL_i(\pi) = \phi(x)\delta(x,y^i;\pi,\mu)^2/(2\eta)$.
Its gradient can be decomposed into token-level contributions  
$
    \na\calL_i(\pi)
    = \sum_{t=1}^{|y^i|} \na\calL_{i,t}(\pi).  
$
Throughout this section, $\na$ denotes the gradient with respect to the parameters of $\pi$ with $\mu$ fixed.

\vspace{0.2cm}
\noindent$\bullet$~Step~1: linearized approximation. 
We linearize the squared-residual loss with respect to the residual $\delta$, around its value at $\pi=\mu$.
Since $\delta(x,y^i;\mu,\mu) = \eta\pr{R(x,y^i)-V^\mu(x)}$,
this replaces the residual factor $\delta(x,y^i;\pi,\mu)/\eta$ in the gradient by $R(x,y^i)-V^\mu(x)$.
For response $y^i$ and the corresponding state $s_t^i = (x, y_{<t}^i)$, the per-token gradient contribution is approximated by  
\eql{\label{eq:est1}}{
    \na \calL_{i,t}(\pi) 
    &= -\phi(x)\frac{\delta(x, y^i; \pi, \mu)}{\eta}\na\prB{\log\pi(y_t^i | s_t^i) + \KL{\mu(\cdot | s_t^i)}{\pi(\cdot | s_t^i)}} \\  
    &\approx -\phi(x)\prB{R(x, y^i) - V^\mu(x)}\na\prB{\log\pi(y_t^i | s_t^i) + \KL{\mu(\cdot | s_t^i)}{\pi(\cdot | s_t^i)}}.
}

\noindent$\bullet$~Step~2: group-based estimation and normalization.
The initial value function $V^\mu(x) = \Eb_{y\sim\cmu(\cdot | x)}\br{R(x, y)}$
has the unbiased estimator $\texttt{mean}\prb{\{R(x, y^j)\}_{j=1}^{G}}$. 
We further choose $\phi(x)$ as the inverse reward standard deviation, i.e., $\phi(x) = \frac{1}{\sqrt{\text{Var}_{y\sim \cmu(\cdot|x)}(R(x, y))}}$. We replace it by its empirical counterpart $\phi(x) \approx \frac{1}{\texttt{std}\prb{\{R(x, y^i)\}_{i=1}^G}}$. 
With these substitutions,~\eqref{eq:est1} becomes
\eql{\label{eq:est2}}{
    \na \calL_{i,t}(\pi) \approx 
    - \adv^i \na\prB{\log\pi(y_t^i | s_t^i) + \KL{\mu(\cdot | s_t^i)}{\pi(\cdot | s_t^i)}}.
}

\noindent$\bullet$~Step~3: binary KL approximation and additive smoothing.  
Computing the full reverse KL divergence in~\eqref{eq:est2} can be costly. 
We therefore replace the full KL divergence $\KL{\mu(\cdot | s_t^i)}{\pi(\cdot | s_t^i)}$ in~\eqref{eq:est2}
with the binary KL divergence $D^{\mathrm{bin}}_{\mathrm{KL}}\pr{\mu(\cdot | s_t^i) \,\Vert\, \pi(\cdot | s_t^i); y_t^i}$  as defined in Section~\ref{sec:binary_kl}.
As proved in Appendix~\ref{sec:proof-binary-kl-gradient}, the following identity holds
\eql{\label{eq:binary-kl-gradient1}}{
    \na\prB{\log\pi(y_t^i | s_t^i) + D^{\mathrm{bin}}_{\mathrm{KL}}\pr{\mu(\cdot | s_t^i) \,\Vert\, \pi(\cdot | s_t^i); y_t^i}} = {\frac{1 - \mu(y_t^i | s_t^i)}{1 - \pi(y_t^i | s_t^i)}}\na\log\pi(y_t^i | s_t^i).
}
Substituting~\eqref{eq:binary-kl-gradient1} into~\eqref{eq:est2} gives
\eql{\label{eq:est3}}{
    \na \calL_{i,t}(\pi) \approx  - \adv^i{\frac{1 - \mu(y_t^i | s_t^i)}{1 - \pi(y_t^i | s_t^i)}}\na \log\pi(y_t^i | s_t^i).
}
The multiplier $\frac{1 - \mu(y_t^i | s_t^i)}{1 - \pi(y_t^i | s_t^i)}$ in~\eqref{eq:est3} can become large when $\pi(y_t^i | s_t^i)$ approaches one. We use the additively smoothed mismatch-correction weight $\wt_t^i$ defined in~\eqref{eq:wti1} for numerical stability.   
The resulting per-token gradient is
\eql{\label{eq:est4}}{
    \na \calL_{i,t}(\pi) 
    \approx -\adv^i\wt_t^i\na\log\pi(y_t^i | s_t^i).
}

\noindent$\bullet$~Step~4: masking and clipping.
The preceding steps yield the approximate per-token gradient direction $-\adv^i\wt_t^i\na\log\pi(y_t^i | s_t^i).$

We apply the GRPO-style clipping mask $M_t^i$ and cap $\wt_t^i$ at $C$. The resulting per-token BPO loss gradient is
\eq{
    \na \calL_{i,t}^{\mathrm{BPO}}(\pi) = -\adv^i M_t^i\min\dr{{\wt_t^i}, C}\na \log\pi(y_t^i | s_t^i).  
}
This gives the BPO loss~\eqref{eq:bpoloss1}.

\section{Experiments}\label{sec:experiments}

\subsection{Experimental Setup}
We evaluate BPO on mathematical reasoning with Qwen3-30B-A3B-Base~\cite{yang2025qwen3}. All methods are trained on the English subset of DAPO-Math-17k~\cite{yu2026dapo}. We compare BPO with GRPO-ClipHigher~\cite{shao2024deepseekmath,yu2026dapo}, GSPO~\cite{zheng2025gspo}, CISPO~\cite{chen2025minimax}, and DPPO~\cite{qi2026rethinking}. We conduct controlled comparisons by varying only the policy loss and keeping all other experimental settings identical. 

Each rollout batch contains 256 prompts with 16 responses per prompt. The resulting 4096 responses are split into 8 minibatches of 512 responses. One training step consists of one rollout batch followed by its eight minibatch updates. We train each method for 400 training steps, corresponding to 3200 optimizer updates. The maximum response length is 16384 tokens. All methods use rollout-router replay (R3)~\cite{ma2025stabilizing}. BPO uses $\epsilon=0.1$ and $C=3.0$. Ablations on these hyperparameters are presented in Appendix~\ref{sec:ablations}. 
More details are given in Appendix~\ref{sec:experimentdetails}.

We evaluate on AIME24, AIME25, and AIME26~\cite{aimeproblems}. We estimate Pass@1 using Avg@32: we sample 32 responses per question, average their correctness, and then average over questions. The average accuracy is the arithmetic mean of the three benchmark scores. Table~\ref{tab:mainaime} reports each method's results at the checkpoint with the highest mean accuracy across the three benchmarks. 

\subsection{Main Results}
Table~\ref{tab:mainaime} shows that BPO achieves 50.5\% average accuracy, compared with 39.5\% for GRPO-ClipHigher and 47.4\% for CISPO, the strongest baseline. These correspond to gains of 11.0 and 3.1 percentage points, respectively. BPO also achieves the highest accuracy on all three benchmarks.

BPO achieves the highest final accuracy (Figure~\ref{fig:mainaime}). After 400 training steps, its average accuracy is 49.4\%, compared with 45.5\% for DPPO, the strongest baseline at the end of training.

\begin{table}[htbp]
\centering
\caption{AIME Avg@32 (\%) on Qwen3-30B-A3B-Base. Each row reports scores at the checkpoint with the highest mean accuracy across the three benchmarks. The best result in each column is bold.}
\label{tab:mainaime}
\small
\renewcommand{\arraystretch}{1.1}
\begin{tabular}{lcccc}
\hline
Method & AIME24 & AIME25 & AIME26 & Avg. \\
\hline
GRPO-ClipHigher & 45.6 & 34.8 & 38.0 & 39.5 \\
GSPO & 50.3 & 35.5 & 44.6 & 43.5 \\
CISPO & 52.7 & 39.0 & 50.4 & 47.4 \\
DPPO & 55.8 & 39.2 & 44.2 & 46.4 \\
BPO & \textbf{57.4} & \textbf{41.0} & \textbf{53.0} & \textbf{50.5} \\
\hline
\end{tabular}
\end{table}

\FloatBarrier

\section{Conclusion}\label{sec:conclusion}
We introduced Bellman Policy Optimization (BPO), a critic-free method for RLVR derived from Policy Mirror Descent. Using the Bellman equations, we obtained a trajectory-level objective that avoids estimating intermediate state values. We proved that this objective and the original PMD objective have the same unique optimal solution on states reachable under the rollout policy. We then derived a practical token-level loss through approximations, with a smoothed mismatch-correction weight replacing the importance-sampling ratio. On Qwen3-30B-A3B-Base, BPO achieves a peak average accuracy of $50.5\%$ across AIME 2024--2026, outperforming GRPO-ClipHigher, GSPO, CISPO, and DPPO by $3.1$--$11.0$ percentage points. Ablations on Qwen3-4B-Base show similar performance across a range of smoothing and truncation settings.

\section*{Acknowledgements}
We thank Sam, Lingfeng, Chris, Simon, Beibin, Yifan, Charlotte, Ziven, Robert, Yuzhen, Yingying, Xingxuan, Zhenwen, Feng, Kaiyu, Kevin, Chen, Chenchen, Ziran, and Marcus for helpful discussions. ChatGPT and Claude were used to edit and proofread the language of this manuscript. Codex and Claude Code were used to assist with debugging code.  

\FloatBarrier

\bibliography{ref}

\begin{thebibliography}{10}

\bibitem{shao2024deepseekmath}
Zhihong Shao, Peiyi Wang, Qihao Zhu, Runxin Xu, Junxiao Song, Xiao Bi, Haowei
  Zhang, Mingchuan Zhang, YK~Li, Yang Wu, et~al.
\newblock Deepseekmath: Pushing the limits of mathematical reasoning in open
  language models.
\newblock {\em arXiv preprint arXiv:2402.03300}, 2024.

\bibitem{guo2025deepseek}
Daya Guo, Dejian Yang, Haowei Zhang, Junxiao Song, Peiyi Wang, Qihao Zhu,
  Runxin Xu, Ruoyu Zhang, Shirong Ma, Xiao Bi, et~al.
\newblock Deepseek-r1: Incentivizing reasoning capability in llms via
  reinforcement learning.
\newblock {\em arXiv preprint arXiv:2501.12948}, 2025.

\bibitem{yang2025qwen3}
An~Yang, Anfeng Li, Baosong Yang, Beichen Zhang, Binyuan Hui, Bo~Zheng, Bowen
  Yu, Chang Gao, Chengen Huang, Chenxu Lv, et~al.
\newblock Qwen3 technical report.
\newblock {\em arXiv preprint arXiv:2505.09388}, 2025.

\bibitem{yu2026dapo}
Qiying Yu, Zheng Zhang, Ruofei Zhu, Yufeng Yuan, Xiaochen Zuo, Yu~Yue, Weinan
  Dai, Tiantian Fan, Gaohong Liu, Lingjun Liu, et~al.
\newblock Dapo: An open-source llm reinforcement learning system at scale.
\newblock {\em Advances in Neural Information Processing Systems},
  38:113222--113244, 2026.

\bibitem{abdin2025phi}
Marah Abdin, Sahaj Agarwal, Ahmed Awadallah, Vidhisha Balachandran, Harkirat
  Behl, Lingjiao Chen, Gustavo de~Rosa, Suriya Gunasekar, Mojan Javaheripi,
  Neel Joshi, et~al.
\newblock Phi-4-reasoning technical report.
\newblock {\em arXiv preprint arXiv:2504.21318}, 2025.

\bibitem{wen2026reinforcement}
Xumeng Wen, Zihan Liu, Shun Zheng, Shengyu Ye, Zhirong Wu, Yang Wang, Zhijian
  Xu, Xiao Liang, Junjie Li, Ziming Miao, et~al.
\newblock Reinforcement learning with verifiable rewards implicitly
  incentivizes correct reasoning in base llms.
\newblock In {\em International Conference on Learning Representations}, volume
  2026, pages 49450--49483, 2026.

\bibitem{zheng2025gspo}
Chujie Zheng, Shixuan Liu, Mingze Li, Xiong-Hui Chen, Bowen Yu, Chang Gao, Kai
  Dang, Yuqiong Liu, Rui Men, An~Yang, Jingren Zhou, and Junyang Lin.
\newblock Group sequence policy optimization.
\newblock {\em arXiv preprint arXiv:2507.18071}, 2025.

\bibitem{qi2026rethinking}
Penghui Qi, Xiangxin Zhou, Zichen Liu, Tianyu Pang, Chao Du, Min Lin, and
  Wee~Sun Lee.
\newblock Rethinking the trust region in llm reinforcement learning.
\newblock {\em arXiv preprint arXiv:2602.04879}, 2026.

\bibitem{ma2026fipo}
Chiyu Ma, Shuo Yang, Kexin Huang, Jinda Lu, Haoming Meng, Shangshang Wang,
  Bolin Ding, Soroush Vosoughi, Guoyin Wang, and Jingren Zhou.
\newblock Fipo: Eliciting deep reasoning with future-kl influenced policy
  optimization.
\newblock {\em arXiv preprint arXiv:2603.19835}, 2026.

\bibitem{mao2026beyond}
Renjie Mao, Xiangxin Zhou, Lvfang Tao, Yixin Ding, Yu~Shi, Yongguang Lin,
  Yuheng Wu, Honglin Zhu, Qian Qiu, and Wenxi Zhu.
\newblock Beyond uniform token-level trust region in llm reinforcement
  learning.
\newblock {\em arXiv preprint arXiv:2606.10968}, 2026.

\bibitem{kang2026vimpo}
Zhewei Kang, Aosong Feng, Sergey Levine, Dawn Song, and Xuandong Zhao.
\newblock Vimpo: Value-implicit policy optimization for llms.
\newblock {\em arXiv preprint arXiv:2606.20008}, 2026.

\bibitem{liu2025understanding}
Zichen Liu, Changyu Chen, Wenjun Li, Penghui Qi, Tianyu Pang, Chao Du, Wee~Sun
  Lee, and Min Lin.
\newblock Understanding r1-zero-like training: A critical perspective.
\newblock {\em arXiv preprint arXiv:2503.20783}, 2025.

\bibitem{chen2025minimax}
Aili Chen, Aonian Li, Bangwei Gong, Binyang Jiang, Bo~Fei, Bo~Yang, Boji Shan,
  Changqing Yu, Chao Wang, Cheng Zhu, et~al.
\newblock Minimax-m1: Scaling test-time compute efficiently with lightning
  attention.
\newblock {\em arXiv preprint arXiv:2506.13585}, 2025.

\bibitem{zhang2026design}
Yifan Zhang, Yifeng Liu, Rina Hughes, Yang Yuan, Quanquan Gu, and Andrew Yao.
\newblock On the design of kl-regularized policy gradient algorithms for llm
  reasoning.
\newblock In {\em International Conference on Learning Representations}, volume
  2026, pages 88357--88395, 2026.

\bibitem{liu2026prorl}
Mingjie Liu, Shizhe Diao, Ximing Lu, Jian Hu, Xin Dong, Yejin Choi, Jan Kautz,
  and Yi~Dong.
\newblock Prorl: Prolonged reinforcement learning expands reasoning boundaries
  in large language models.
\newblock {\em Advances in Neural Information Processing Systems},
  38:17998--18031, 2026.

\bibitem{schulman2017proximal}
John Schulman, Filip Wolski, Prafulla Dhariwal, Alec Radford, and Oleg Klimov.
\newblock Proximal policy optimization algorithms.
\newblock {\em arXiv preprint arXiv:1707.06347}, 2017.

\bibitem{blakeman2025nvidia}
Aaron Blakeman, Aaron Grattafiori, Aarti Basant, Abhibha Gupta, Abhinav
  Khattar, Adi Renduchintala, Aditya Vavre, Akanksha Shukla, Akhiad Bercovich,
  Aleksander Ficek, et~al.
\newblock Nvidia nemotron 3: Efficient and open intelligence.
\newblock {\em arXiv preprint arXiv:2512.20856}, 2025.

\bibitem{blakeman2026nemotron}
Aaron Blakeman, Aaron Thomas, Aastha Jhunjhunwala, Abhibha Gupta, Abhinav
  Khattar, Adam Rajfer, Adi Renduchintala, Adil Asif, Aditya Vavre,
  Adriana~Flores Miranda, et~al.
\newblock Nemotron 3 ultra: Open, efficient mixture-of-experts hybrid
  mamba-transformer model for agentic reasoning.
\newblock {\em arXiv preprint arXiv:2606.15007}, 2026.

\bibitem{team2025every}
Ling Team, Anqi Shen, Baihui Li, Bin Hu, Bin Jing, Cai Chen, Chao Huang, Chao
  Zhang, Chaokun Yang, Cheng Lin, et~al.
\newblock Every step evolves: Scaling reinforcement learning for trillion-scale
  thinking model.
\newblock {\em arXiv preprint arXiv:2510.18855}, 2025.

\bibitem{lan2023policy}
Guanghui Lan.
\newblock Policy mirror descent for reinforcement learning: Linear convergence,
  new sampling complexity, and generalized problem classes.
\newblock {\em Mathematical programming}, 198(1):1059--1106, 2023.

\bibitem{xiao2022convergence}
Lin Xiao.
\newblock On the convergence rates of policy gradient methods.
\newblock {\em Journal of Machine Learning Research}, 23(282):1--36, 2022.

\bibitem{pmlr-v267-kazemnejad25a}
Amirhossein Kazemnejad, Milad Aghajohari, Eva Portelance, Alessandro Sordoni,
  Siva Reddy, Aaron Courville, and Nicolas Le~Roux.
\newblock {V}ine{PPO}: Refining credit assignment in {RL} training of {LLM}s.
\newblock In Aarti Singh, Maryam Fazel, Daniel Hsu, Simon Lacoste-Julien, Felix
  Berkenkamp, Tegan Maharaj, Kiri Wagstaff, and Jerry Zhu, editors, {\em
  Proceedings of the 42nd International Conference on Machine Learning}, volume
  267 of {\em Proceedings of Machine Learning Research}, pages 29557--29590.
  PMLR, 13--19 Jul 2025.

\bibitem{rafailov2023direct}
Rafael Rafailov, Archit Sharma, Eric Mitchell, Stefano Ermon, Christopher~D
  Manning, and Chelsea Finn.
\newblock Direct preference optimization: your language model is secretly a
  reward model.
\newblock In {\em Proceedings of the 37th International Conference on Neural
  Information Processing Systems}, pages 53728--53741, 2023.

\bibitem{zeng2024token}
Yongcheng Zeng, Guoqing Liu, Weiyu Ma, Ning Yang, Haifeng Zhang, and Jun Wang.
\newblock Token-level direct preference optimization.
\newblock In {\em International Conference on Machine Learning}, pages
  58348--58365. PMLR, 2024.

\bibitem{rafailovr}
Rafael Rafailov, Joey Hejna, Ryan Park, and Chelsea Finn.
\newblock From $ r $ to $ q^* $: Your language model is secretly a q-function.
\newblock In {\em First Conference on Language Modeling}.

\bibitem{aimeproblems}
{Art of Problem Solving}.
\newblock {AIME} problems and solutions.

\bibitem{zheng2025stabilizing}
Chujie Zheng, Kai Dang, Bowen Yu, Mingze Li, Huiqiang Jiang, Junrong Lin,
  Yuqiong Liu, Hao Lin, Chencan Wu, Feng Hu, et~al.
\newblock Stabilizing reinforcement learning with llms: Formulation and
  practices.
\newblock {\em arXiv preprint arXiv:2512.01374}, 2025.

\bibitem{qi2025defeating}
Penghui Qi, Zichen Liu, Xiangxin Zhou, Tianyu Pang, Chao Du, Wee~Sun Lee, and
  Min Lin.
\newblock Defeating the training-inference mismatch via fp16.
\newblock {\em arXiv preprint arXiv:2510.26788}, 2025.

\bibitem{yao2026rethinking}
Jiarui Yao, Xiangxin Zhou, Penghui Qi, Wee~Sun Lee, Liefeng Bo, and Tianyu
  Pang.
\newblock Rethinking the divergence regularization in llm rl.
\newblock {\em arXiv preprint arXiv:2606.09821}, 2026.

\bibitem{yao2025rollout}
Feng Yao, Liyuan Liu, Dinghuai Zhang, Chengyu Dong, Jingbo Shang, and Jianfeng
  Gao.
\newblock On the rollout-training mismatch in modern rl systems.
\newblock In {\em OPT 2025: Optimization for Machine Learning}, 2025.

\bibitem{sutton1998reinforcement}
Richard~S Sutton, Andrew~G Barto, and Andrew Barto.
\newblock {\em Reinforcement learning: An introduction}, volume~1.
\newblock MIT press Cambridge, 1998.

\bibitem{geist2019theory}
Matthieu Geist, Bruno Scherrer, and Olivier Pietquin.
\newblock A theory of regularized markov decision processes.
\newblock In {\em International conference on machine learning}, pages
  2160--2169. PMLR, 2019.

\bibitem{tomar2020mirror}
Manan Tomar, Lior Shani, Yonathan Efroni, and Mohammad Ghavamzadeh.
\newblock Mirror descent policy optimization.
\newblock {\em arXiv preprint arXiv:2005.09814}, 2020.

\bibitem{ma2025stabilizing}
Wenhan Ma, Hailin Zhang, Liang Zhao, Yifan Song, Yudong Wang, Zhifang Sui, and
  Fuli Luo.
\newblock Stabilizing moe reinforcement learning by aligning training and
  inference routers.
\newblock {\em arXiv preprint arXiv:2510.11370}, 2025.

\bibitem{ivison2026tmax}
Hamish Ivison, Junjie~Oscar Yin, Rulin Shao, Teng Xiao, Nathan Lambert, and
  Hannaneh Hajishirzi.
\newblock Tmax: A simple recipe for terminal agents.
\newblock {\em arXiv preprint arXiv:2606.23321}, 2026.

\bibitem{khatri2026art}
Devvrit Khatri, Lovish Madaan, Rishabh Tiwari, Rachit Bansal, Venkata Sai
  Surya~Subramanyam Duvvuri, Manzil Zaheer, Inderjit Dhillon, David
  Brandfonbrener, and Rishabh Agarwal.
\newblock The art of scaling reinforcement learning compute for llms.
\newblock In {\em International Conference on Learning Representations}, volume
  2026, pages 72438--72467, 2026.

\end{thebibliography}

\begin{appendices}

\section{Omitted Proofs}
\subsection{Proof of Theorem~\ref{thm:delta2}}\label{sec:prfdelta2}
Sufficiency follows from~\eqref{eq:sumdelta2}.
It remains to prove necessity. 
We use $\mathbb{I}_A$ to denote the indicator of an event $A$ throughout this section.  

Let $\text{supp}\mu(\cdot|s)$ denote the set of actions with positive probability under policy $\mu$ at state $s$.
Let $\text{supp}\cmu(\cdot|x)$ denote the set of responses with positive probability under policy $\mu$ given prompt $x$.
Let $\calS_x^\mu$ denote the set of states reachable from prompt $x$ with positive probability under policy $\mu$.

Fix any prompt $x\in D$.
For any state-action pair $(s,a)$, define
\eq{
    d_\pi(s,a) = \eta A^\mu(s,a) - \prB{\log\pi(a|s) - \log\mu(a|s) + \KL{\mu(\cdot|s)}{\pi(\cdot|s)}}.
}
By definition, 
\eql{\label{eq:deltasumdt}}{
    \delta(x,y;\pi,\mu) = \sum_{t=1}^{|y|} d_\pi(s_t,y_t).
}
We first show that the minimum of~\eqref{eq:delta2} is zero.
The PMD optimality condition~\eqref{eq:PMDKL1} gives
\eq{
    d_{\pi^+}(s,a)=0.
}
The exponential-form update~\eqref{eq:piexpwadv1} preserves the support of the rollout policy, so $\pi^+\in\Pi_\mu$ and
\eq{
    \delta(x,y;\pi^+,\mu)=0,
    \quad
    \forall y\in\text{supp}\cmu(\cdot|x).
}
Hence, $\pi^+$ is feasible for~\eqref{eq:delta2} and achieves objective value zero.
Since the objective is nonnegative, the minimum value of~\eqref{eq:delta2} is zero.

Now let $\hat{\pi}^+\in\Pi_\mu$ be any optimal solution
of~\eqref{eq:delta2}.
Since the minimum value is zero and $\phi(x) > 0$, we have  
\eql{\label{eq:deltapizero1}}{
    \delta(x,Y;\hat{\pi}^+,\mu)=0,
    \quad
    Y\sim\cmu(\cdot|x)\text{-almost surely}.  
}
Since the objective value is finite, for every state $s\in\calS_x^\mu$,
$\KL{\mu(\cdot|s)}{\hat{\pi}^+(\cdot|s)} < \infty. $
Consequently, $\mu(\cdot|s)$ is absolutely continuous with respect to $\hat{\pi}^+(\cdot|s)$ for any $s\in \calS_x^{\mu}$. 
Thus, all quantities below are finite on states reachable under $\mu$.  

Let $\tau$ be the time at which generation terminates:  
\eq{
    \tau = \inf\dr{t\geq 1: Y_t=\mathrm{EOS}} \wedge T_{\max}.
}
Define the filtration
\eq{
    \calF_t = \sigma\pr{Y_1,\dots,Y_{t\wedge\tau}}, \quad 0\leq t\leq T_{\max}.   
}
For $1\leq t\leq T_{\max}$, define
\eq{
    D_t = \mathbb{I}_{\dr{t\leq\tau}}d_{\hat{\pi}^+}(S_t,Y_t), \quad
    M_n = \sum_{t=1}^n D_t,
}
where $S_t=(x,Y_{<t})$ on the event $\dr{t\leq\tau}$.

Since $\mathbb{I}_{\dr{t\leq\tau}}$ and $S_t$ are
$\calF_{t-1}$-measurable, we have
\eq{
    &\quad \Eb\br{D_t\mid\calF_{t-1}} \\
    &= \mathbb{I}_{\dr{t\leq\tau}}\sum_{a\in\mathcal V}\mu(a|S_t)d_{\hat{\pi}^+}(S_t,a)
    \\
    &= \mathbb{I}_{\dr{t\leq\tau}}\eta\sum_{a\in\mathcal V}\mu(a|S_t)A^\mu(S_t,a)
    \\
    &\quad
    - \mathbb{I}_{\dr{t\leq\tau}}\prB{\sum_{a\in\mathcal V}\mu(a|S_t)\log\frac{\hat{\pi}^+(a|S_t)}{\mu(a|S_t)}
    +
    \KL{\mu(\cdot|S_t)}{\hat{\pi}^+(\cdot|S_t)}}
    \\
    &=0,  
}
where the last equality follows from~\eqref{eq:avgA1}. 
Thus, $\dr{M_n}_{n=0}^{T_{\max}}$ is an integrable martingale.

By~\eqref{eq:deltasumdt} and~\eqref{eq:deltapizero1},
\eq{
    M_{T_{\max}} = \sum_{t=1}^{\tau} d_{\hat{\pi}^+}(S_t,Y_t) = \delta(x,Y;\hat{\pi}^+,\mu) = 0, \quad \text{$\cmu(\cdot|x)$-almost surely.}
}
Thus, for any $0\leq n\leq T_{\max}$,
\eq{
    M_n = \Eb\br{M_{T_{\max}} | \calF_n} = 0,
    \quad \text{$\cmu(\cdot|x)$-almost surely.}
}
It follows that for any $1\leq t\leq T_{\max}$,  
\eq{
    D_t = M_t - M_{t-1} = 0 \quad \text{$\cmu(\cdot|x)$-almost surely.}
}
Consider any state $s\in\calS_x^\mu$ and any action
$a\in\text{supp}\mu(\cdot|s)$.
For the corresponding step $t$, the event $\dr{S_t=s, Y_t=a}$ has strictly positive probability under $\cmu(\cdot|x)$. 
Since $D_t=0$ almost surely, this implies
\eq{
    d_{\hat{\pi}^+}(s,a)=0.
}
Combining this with $d_{\pi^+}(s,a)=0$, we obtain
\eq{
    \log\frac{\hat{\pi}^+(a|s)}{\pi^+(a|s)} = \KL{\mu(\cdot|s)}{\pi^+(\cdot|s)} - \KL{\mu(\cdot|s)}{\hat{\pi}^+(\cdot|s)}.  
}
The right-hand side does not depend on $a$.
Hence, for every $s\in\calS_x^\mu$, there exists a constant $c(s)>0$ such that
\eq{
    \hat{\pi}^+(a|s) = c(s)\pi^+(a|s), \quad \forall a\in\text{supp}\mu(\cdot|s).
}
Since $\hat{\pi}^+,\pi^+\in\Pi_\mu$, both policies assign zero probability outside $\text{supp}\mu(\cdot|s)$.
Therefore,  
\eq{
    1 = \sum_{a\in\text{supp}\mu(\cdot|s)}
    \hat{\pi}^+(a|s)
    = c(s)\sum_{a\in\text{supp}\mu(\cdot|s)}
    \pi^+(a|s)
    = c(s).  
}
Thus, $c(s)=1$, and
\eq{
    \hat{\pi}^+(\cdot|s) = \pi^+(\cdot|s), \quad \forall s\in\calS_x^\mu.
}
In addition, since $\hat{\pi}^+,\pi^+\in\Pi_\mu$, both $\pi^+$ and $\hat{\pi}^+$ assign zero probability outside $\text{supp}\cmu(\cdot|x)$.
Consequently,
\eq{
    \cpx{\hat{\pi}^+}(\cdot|x) = \cpx{\pi^+}(\cdot|x).  
}
This completes the proof.
\hfill $\square$

\subsection{Proof of the Identity~(\ref{eq:binary-kl-gradient1})}\label{sec:proof-binary-kl-gradient}
Fix a state $s$ and an action $y$, and let
\eq{
    p = \pi(y | s),\quad q = \mu(y | s).
}
The binary KL divergence associated with action $y$ is  
\eq{
    D^{\mathrm{bin}}_{\mathrm{KL}}\pr{\mu(\cdot | s) \,\Vert\, \pi(\cdot | s); y} = q\log\frac{q}{p} + (1 - q)\log\frac{1 - q}{1 - p}.
}
Differentiating with respect to $\pi$ gives
\eq{
    \na D^{\mathrm{bin}}_{\mathrm{KL}}\pr{\mu(\cdot | s) \,\Vert\, \pi(\cdot | s); y} = -q\na\log p + \frac{1 - q}{1 - p}\na p.
}
Using the fact that $\na p = p\na\log p$, we obtain
\eq{
    \na\prB{\log p + D^{\mathrm{bin}}_{\mathrm{KL}}\pr{\mu(\cdot | s) \,\Vert\, \pi(\cdot | s); y}} 
    = (1 - q)\na\log p + \frac{(1 - q)p}{1 - p}\na\log p
    = \frac{1 - \mu(y | s)}{1 - \pi(y | s)}\na\log\pi(y | s),                 
}
i.e.,  
\eq{
    \na\prB{\log\pi(y | s) + D^{\mathrm{bin}}_{\mathrm{KL}}\pr{\mu(\cdot | s) \,\Vert\, \pi(\cdot | s); y}} = \frac{1 - \mu(y | s)}{1 - \pi(y | s)}\na\log\pi(y | s).
}
This proves~\eqref{eq:binary-kl-gradient1}.  
\hfill $\square$

\section{Experimental Details}\label{sec:experimentdetails}

All experiments use AdamW with a constant learning rate of $10^{-6}$, $\beta_1=0.9$, $\beta_2=0.98$, and weight decay $0.1$. The gradient clipping threshold is $1.0$. We use no KL penalty and the entropy bonus coefficient is zero. 
Training rollouts are sampled at temperature $1.0$ and top-$p=1.0$, with top-$k$ filtering disabled. Advantages are centered and normalized by the reward standard deviation within each response group as in GRPO. 
We use \texttt{seq-mean-token-mean} for loss aggregation. 
These settings are shared across methods.

We use 400 training steps for the main experiments in Section~\ref{sec:experiments} and 1000 for the ablations in Appendix~\ref{sec:ablations}. 
The AIME benchmarks are available on Hugging Face.\footnote{AIME24: \url{https://huggingface.co/datasets/math-ai/aime24}, AIME25: \url{https://huggingface.co/datasets/math-ai/aime25}, AIME26: \url{https://huggingface.co/datasets/math-ai/aime26}.} We evaluate every ten rollout batches (i.e., training steps). 
All evaluations use 32 responses per question at temperature $1.0$ and top-$p=1.0$, with top-$k$ filtering disabled. Evaluation uses the same response-length limit as training for each model.

We report Avg@32 as an estimate of Pass@1. 
The Avg. column and the plotted average accuracy are the arithmetic mean of the AIME24, AIME25, and AIME26 Avg@32 scores.

We select the checkpoint with the highest mean Avg@32 across the three benchmarks, choosing the earlier step in a tie. 

GRPO-ClipHigher uses the asymmetric clipping interval $[0.8,1.28]$ from DAPO~\cite{yu2026dapo}. CISPO uses an importance-weight cap of $3.0$, following the configuration used in the stability experiments of DPPO~\cite{qi2026rethinking}. DPPO uses binary total variation with threshold $\delta=0.1$, following TMax~\cite{ivison2026tmax}. GSPO uses the clipping interval $[1-3\times10^{-3},1+5\times10^{-3}]$, following the ablation results in~\cite{khatri2026art}. BPO uses the default setting $\epsilon=0.1$ and $C=3.0$ in the main experiments.

\section{Ablation Studies}\label{sec:ablations}
We study the two BPO hyperparameters $\eps$ and $C$ on Qwen3-4B-Base with 1000 training steps. 
Here, each batch contains 128 prompts with 16 responses per prompt. The 2048 responses are split into four minibatches of 512, with a maximum response length of 8192 tokens. One training step in the figures denotes one rollout batch followed by its minibatch updates.
Each sweep changes only one hyperparameter while keeping the other settings fixed. Both sweeps in Section~\ref{sec:epsilonablation} and Section~\ref{sec:capablation} use the training setup in Appendix~\ref{sec:experimentdetails} and share the same GRPO-ClipHigher baseline.

\subsection{Sensitivity to $\epsilon$}\label{sec:epsilonablation}
\begin{figure}[htbp]
\centering
\includegraphics[width=0.8\linewidth]{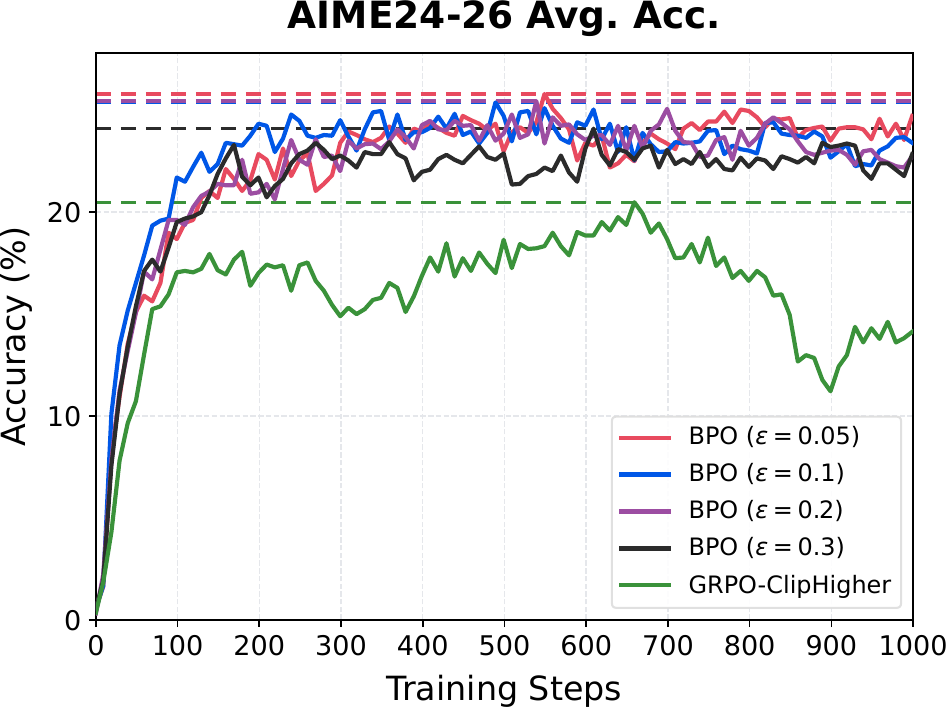}
\caption{Effect of BPO's smoothing parameter $\epsilon$ on Qwen3-4B-Base with $C=3.0$. Curves show mean Avg@32 across AIME24--26. Dashed lines mark the peak of each curve.}
\label{fig:epsilonablation}
\end{figure}

We fix $C=3.0$ and vary $\epsilon$ over $\{0.05,0.1,0.2,0.3\}$. Figure~\ref{fig:epsilonablation} shows the training curves, and Table~\ref{tab:epsilonablation} summarizes the results.

\begin{table}[htbp]
\centering
\caption{Ablation on BPO's smoothing parameter $\epsilon$ with $C=3.0$, reporting AIME Avg@32 (\%). Each row reports scores at the checkpoint with the highest mean accuracy across the three benchmarks. The best result in each column is bold.}
\label{tab:epsilonablation}
\small
\renewcommand{\arraystretch}{1.1}
\begin{tabular}{lrrrr}
\hline
Method & AIME24 & AIME25 & AIME26 & Avg. \\
\hline
BPO ($\epsilon=0.05$) & \textbf{31.6} & 23.0 & 22.8 & \textbf{25.8} \\
BPO ($\epsilon=0.1$) & 27.6 & \textbf{26.8} & 21.8 & 25.4 \\
BPO ($\epsilon=0.2$) & 29.1 & 24.3 & \textbf{23.0} & 25.5 \\
BPO ($\epsilon=0.3$) & 26.4 & 25.5 & 20.4 & 24.1 \\
GRPO-ClipHigher & 22.1 & 22.8 & 16.6 & 20.5 \\
\hline
\end{tabular}
\end{table}

BPO achieves similar average accuracies of 25.4\%--25.8\% for $\epsilon\in\{0.05,0.1,0.2\}$. At $\epsilon=0.3$, accuracy is 24.1\%, still 3.6 percentage points above GRPO-ClipHigher. BPO shows little sensitivity to $\epsilon$ from $0.05$ to $0.2$ and outperforms the baseline at all four settings.

\FloatBarrier

\subsection{Sensitivity to $C$}\label{sec:capablation}
\begin{figure}[htbp]
\centering
\includegraphics[width=0.8\linewidth]{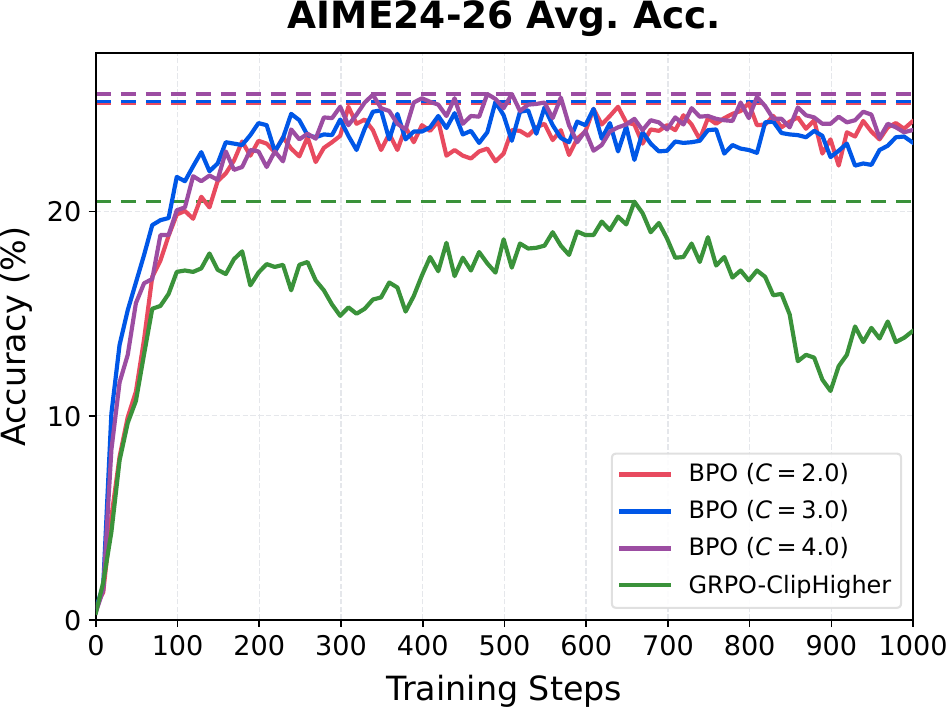}
\caption{Effect of BPO's truncation parameter $C$ on Qwen3-4B-Base with $\epsilon=0.1$. Curves show mean Avg@32 across AIME24--26. Dashed lines mark the peak of each curve.}
\label{fig:capablation}
\end{figure}

We fix $\epsilon=0.1$ and vary $C$ over $\{2.0,3.0,4.0\}$. Figure~\ref{fig:capablation} shows the training curves, and Table~\ref{tab:capablation} summarizes the results.

\begin{table}[htbp]
\centering
\caption{Ablation on BPO's truncation parameter $C$ with $\epsilon=0.1$, reporting AIME Avg@32 (\%). Each row reports scores at the checkpoint with the highest mean accuracy across the three benchmarks. The best result in each column is bold.}
\label{tab:capablation}
\small
\renewcommand{\arraystretch}{1.1}
\begin{tabular}{lrrrr}
\hline
Method & AIME24 & AIME25 & AIME26 & Avg. \\
\hline
BPO ($C=2.0$) & 26.4 & 26.0 & \textbf{23.5} & 25.3 \\
BPO ($C=3.0$) & \textbf{27.6} & 26.8 & 21.8 & 25.4 \\
BPO ($C=4.0$) & 27.1 & \textbf{29.0} & 21.3 & \textbf{25.8} \\
GRPO-ClipHigher & 22.1 & 22.8 & 16.6 & 20.5 \\
\hline
\end{tabular}
\end{table}

The three values of $C$ yield average accuracies between 25.3\% and 25.8\%, a spread of 0.5 percentage points. All three exceed the GRPO-ClipHigher baseline of 20.5\%. These results show that BPO is insensitive to $C$ from $2.0$ to $4.0$.

\end{appendices}

\end{document}